\documentclass{article}

\usepackage{microtype}
\usepackage{graphicx}
\usepackage{subfigure}
\usepackage{booktabs} 

\usepackage{hyperref}

\usepackage[accepted]{icml2025}

\usepackage{amsmath}
\usepackage{amssymb}
\usepackage{mathtools}
\usepackage{amsthm}

\usepackage[capitalize,noabbrev]{cleveref}

\theoremstyle{plain}

\theoremstyle{definition}

\theoremstyle{remark}

\usepackage[textsize=tiny]{todonotes}

\usepackage{multirow}
\usepackage{arydshln}
\usepackage{makecell}
\usepackage{colortbl}
\usepackage{graphicx}
\newcommand{\etal}{\textit{et al.}}
\newcommand{\eg}{\textit{e.g.}}

\begin{document}

\twocolumn[
\icmltitle{Fine-Grained Captioning of Long Videos through Scene Graph Consolidation}


\icmlsetsymbol{equal}{*}

\begin{icmlauthorlist}
\icmlauthor{Sanghyeok Chu}{ece}
\icmlauthor{Seonguk Seo$^\dagger$}{ece}
\icmlauthor{Bohyung Han}{ece,ipai}
\end{icmlauthorlist}

\icmlaffiliation{ece}{ECE, Seoul National University, Korea.}
\icmlaffiliation{ipai}{IPAI, Seoul National University, Korea. $\dagger$: Currently at Meta}
\icmlcorrespondingauthor{Bohyung Han}{bhhan@snu.ac.kr}

\icmlkeywords{Machine Learning, ICML, video captioning, long video captioning}

\vskip 0.3in
]



\printAffiliationsAndNotice{}  


\begin{abstract}

Recent advances in vision-language models have led to impressive progress in caption generation for images and short video clips. 
However, these models remain constrained by their limited temporal receptive fields, making it difficult to produce coherent and comprehensive captions for long videos. 
While several methods have been proposed to aggregate information across video segments, they often rely on supervised fine-tuning or incur significant computational overhead.
To address these challenges, we introduce a novel framework for long video captioning based on graph consolidation. 
Our approach first generates segment-level captions, corresponding to individual frames or short video intervals, using off-the-shelf visual captioning models.
These captions are then parsed into individual scene graphs, which are subsequently consolidated into a unified graph representation that preserves both holistic context and fine-grained details throughout the video. 
A lightweight graph-to-text decoder then produces the final video-level caption. 
This framework effectively extends the temporal understanding capabilities of existing models without requiring any additional fine-tuning on long video datasets. Experimental results show that our method significantly outperforms existing LLM-based consolidation approaches, achieving strong zero-shot performance while substantially reducing computational costs.

\end{abstract}


\section{Introduction}
\label{sec:intro}

Vision-language models (VLMs) have demonstrated impressive capabilities across diverse vision-language tasks, including visual question answering, visual dialogue, cross-modal retrieval, and spatiotemporal understanding~\cite{alayrac2022flamingo, instructblip, gpt4v, chen2024far, huang2024vtimellm, zhang2025videollama, xu2024pllava, Maaz2023VideoChatGPT}.
Notably, substantial progress has been made in generating captions for images and short video clips~\cite{liu2024llavanext, chai2024auroracap, zhao2024videoprism, wang2024internvideo2, chen2024expanding, mun2019streamlined}.

However, generating captions for longer videos remains a significant challenge.
Most existing models are designed for short-term visual inputs, such as images or short video clips, and lack effective support for holistic encoding of entire long videos.
As a result, captioning videos beyond a model's temporal window typically requires processing and integrating information from multiple temporal segments.
Several approaches, such as memory-based~\cite{zhou2024streaming, song2024moviechat, balazevicmemory} and recursive frameworks~\cite{zhou2024streaming, islam2024video, qian2024streaming, weng2024longvlm, kahatapitiya2024language}, have been proposed to consolidate information across these segments.
However, these methods often rely on supervised fine-tuning with the target datasets, which limits their generalizability to unseen video domains.
More recently, large language models (LLMs) have been employed to generate textual summaries across multiple video segments~\cite{wang2022language, chen2023video, zhang2024simple}.
While these LLM-based approaches eliminate the need to adapt existing models for long videos, they typically incur high inference overhead and require significant computational resources.

To address these limitations, we propose a novel framework that integrates segment-level captions into a unified global description via graph-based consolidation.
We first obtain segment-level captions---each corresponding to either a single frame or a short video clip, depending on the chosen visual captioning model---using an off-the-shelf captioning algorithm.
Each caption is then parsed into a scene graph, and these graphs are consolidated into a unified structure that captures the comprehensive semantics of the entire video.
Finally, a lightweight graph-to-text decoder, trained solely on external text corpora, translates the consolidated graph into a coherent global caption.

The proposed approach enhances understanding and processing of long-range temporal information without requiring architectural changes or fine-tuning on long video datasets.
In particular, our framework can be paired with any off-the-shelf VLM, effectively extending its captioning capability beyond the model's inherent temporal constraints.
Unlike other LLM-based consolidation methods, it minimizes computational overhead by employing a lightweight graph-to-text decoder with significantly fewer parameters. 
Our experimental results demonstrate that our approach achieves superior performance in both zero-shot video captioning and zero-shot video paragraph captioning, demonstrating its effectiveness and efficiency.

In summary, our key contributions are organized as follows:
\begin{itemize}
	\item We propose a novel approach to generate fine-grained captions for long videos using the information across multiple temporal segments.
	\item We introduce a graph consolidation algorithm that merges segment-level scene graphs into a unified representation to capture both holistic context and fine-grained details across the entire video.
 	\item Our method achieves strong zero-shot captioning performance with significantly lower computational cost compared to LLM-based approaches.
\end{itemize}


\section{Related Works}
\label{sec:related}

\paragraph{Video captioning}
Recent advances in video captioning have predominantly rely on supervised training using large-scale datasets, achieving impressive results across various benchmarks~\cite{lei2021less, wang2022omnivl, yan2022videococa, liu2024llavanext, zhao2024videoprism, wang2024internvideo2, chen2024expanding}. 
However, extending these supervised approaches to longer videos remains challenging, primarily due to the scarcity of annotated data covering extensive temporal contexts and the computational complexity involved in modeling long-range dependencies. 
While various methods have been proposed to tackle these challenges, the needs for supervised fine-tuning for specific target datasets hampers scalability and generalization to unseen video domains~\cite{yang2023vid2seq, islam2024video, song2024moviechat, balazevicmemory, qian2024streaming, weng2024longvlm, kahatapitiya2024language}.

\paragraph{Zero-shot video captioning}
Researchers have explored methods for video captioning without using paired video-text annotations.
One approach involves refining language model outputs solely at test time.
ZeroCap~\cite{tewel2021zero} and related methods~\cite{Tewel_2023_BMVC} use image-text alignment score calculated by CLIP~\cite{radford2021learning} in gradient updates to adjust language model features, while MAGIC~\cite{su2022language} employs a CLIP-induced decoding strategy to ensure semantic relevance. 
Although initially developed for images, these methods extend to videos by aggregating frame-level features into a single representation.
Another approach, often termed zero-shot, involves text-only training without paired video-text annotations, where text decoders are used in conjunction with image-text aligned encoders such as CLIP and ImageBind~\cite{girdhar2023imagebind}. 
Methods such as DeCap~\cite{lidecap} and C$^{3}$~\cite{zhang2024connect} generate captions by aligning visual and textual features in a shared embedding space.
However, these approaches often fail to produce accurate and coherent captions, especially when applied to videos with complex events.

\paragraph{Zero-shot long video captioning}
Generating coherent and comprehensive captions for long-context videos under zero-shot settings often relies on the consolidation of information derived from multiple temporal segments. 
Existing consolidation techniques, including memory-based~\cite{zhou2024streaming, song2024moviechat, balazevicmemory} and recursive approaches~\cite{islam2024video, qian2024streaming, weng2024longvlm, kahatapitiya2024language}, require supervised fine-tuning on the target dataset, which limits their applicability to zero-shot scenarios.
Recently, LLMs have emerged as a promising tool for zero-shot consolidation, leveraging their general reasoning capabilities without task-specific fine-tuning.
For example, VidIL~\cite{wang2022language} constructs prompts by integrating multi-level textual information from image-language models, including objects, events, attributes, frame captions, and subtitles. 
Due to the complexity of these prompts, it incorporates illustrative few-shot exemplars from training dataset, to guide LLMs in interpreting and utilizing these textual cues for video captioning
Similarly, Video ChatCaptioner~\cite{chen2023video} adopts an interactive framework, where an LLM queries an image VLM for captions of individual frames and aggregates them to generate video caption.
While these LLM-based methods are powerful and flexible, they typically incur high computational costs.


\section{Scene Graph Construction for Videos}
\label{sec:scene}
\begin{figure*}[t]
    \centering
    \includegraphics[width=\textwidth]{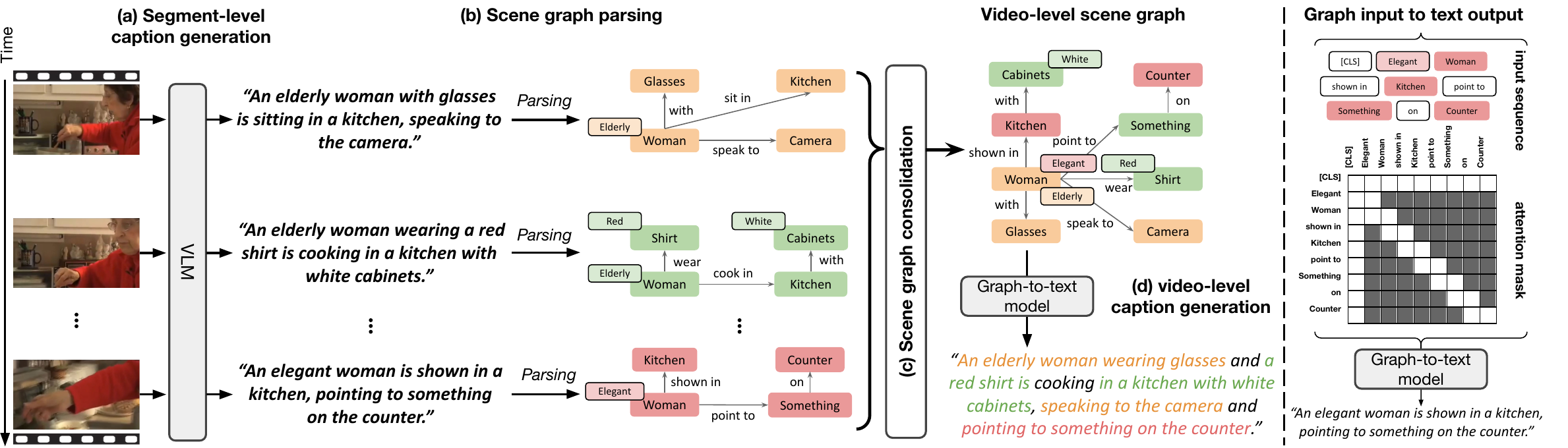}
    \vspace{-5mm}
    \caption{An overview of our zero-shot video caption generation framework. (left): The pipeline consists of (a) segment-level caption generation using off-the-shelf VLMs, (b) scene graph parsing for each caption, (c) consolidation of individual scene graphs into a unified graph representing the entire video, and (d) video caption generation through our graph-to-text model. 
    (right): Illustration of how the scene graph is transformed into an input for the graph-to-text model to generate a caption.
    }
    \label{fig:framework}
\end{figure*}

To enable effective captioning of long videos, we propose a novel framework that constructs and consolidates scene graphs derived from segment-level captions, as illustrated in Figure~\ref{fig:framework}. 
The framework comprises four main stages: (1) generating captions for individual video segments using VLMs, (2) converting these captions into scene graphs, (3) merging the scene graphs from all segments into a unified graph, and (4) generating a comprehensive description from the consolidated graph. 
By aggregating information across segments, the proposed method produces captions that are more coherent and contextually informative, capturing fine-grained details throughout the video.
Throughout this paper, we use the term \textit{segment} to denote a temporal unit of a video---either a single frame or a short interval---depending on the characteristics of the employed VLM.

\subsection{Generating segment-level captions}
Given an input video, we first divide it into a series of temporal segments. 
We then generate segment-level captions using off-the-shelf VLMs, with prompts guiding the models to produce descriptive sentences suitable for scene graph construction.
While we primarily utilize open-source VLMs as our captioning backbone, our framework is flexible enough to incorporate any VLM, including proprietary or closed-source models, as long as APIs are accessible.

\subsection{Parsing captions into scene graphs}
A scene graph $G = (\mathcal{O}, \mathcal{E})$ is defined by a set of objects\, $\mathcal{O} = \{o_1, o_2, \ldots \}$, and a set of edges between objects, $\mathcal{E}$.
Each object $o_i = (c_i, \mathcal{A}_i)$ consists of an object class $c_i \in \mathcal{C}$ and its attribute set $\mathcal{A}_i \subseteq \mathcal{A}$, where $\mathcal{C}$ is a set of object classes and $\mathcal{A}$ is a set of all possible attributes.
A directed edge, $e_{i,j} \equiv (o_i, o_j) \in \mathcal{E}$, has a label $r_{i,j} \in \mathcal{R}$, specifying the relationship from one object to the other.
All object classes, attributes, and relationship labels are represented as text strings.

We convert the generated caption from each segment into a scene graph, providing a more structured understanding of each segment.
A caption is parsed into a scene graph by textual scene graph parser, and FACTUAL-MR parser~\cite{li-etal-2023-factual} is used in our implementation.
This parser first maps the caption to an intermediate semantic representation consisting of objects, attributes, and relationships, then deterministically converts it into a scene graph.
By representing each segment as a graph consisting of objects and their relationships, we can apply a graph merging technique to produce a holistic representation of the entire input video.

\subsection{Scene graph consolidation}
\label{sub:scene}
%

\begin{algorithm}[t]
\caption{Scene graph consolidation}
\label{alg:hierarchical_graph_merge}
\begin{algorithmic}[1]

  \STATE \textbf{Input:} 
  \STATE \quad $\mathcal{G} = \{ G_1, G_2, \dots, G_n \}$: set of scene graphs
  \STATE \quad $\phi(\cdot)$: a graph encoder
  \STATE \quad $\psi_i(\cdot)$: a function returning the $i^\text{th}$ object in a graph
  \STATE \quad $\pi$: a permutation function
  \STATE \quad $\tau$: a threshold

  \STATE \textbf{Output:} $G_{\text{video}}$: a video-level scene graph

  \WHILE{$|\mathcal{G}| > 1$}
    \STATE $\text{Retrieve the most similar pair } \{G^s, G^t \} \text{ from }\mathcal{G}$
    \STATE $G^s = (\mathcal{O}^s, \mathcal{E}^s),\,G^t = (\mathcal{O}^t, \mathcal{E}^t)$
    \STATE $G^m = (\mathcal{O}^m, \mathcal{E}^m) \gets (\mathcal{O}^s \cup \mathcal{O}^t, \mathcal{E}^s \cup \mathcal{E}^t)$
    \STATE $\pi^* \gets \displaystyle \arg\max_{\pi \in \Pi} \sum_{i} 
      \frac{\psi_i(\phi(G^s))}{\lVert \psi_i(\phi(G^s)) \rVert} \; \cdot \;
      \frac{\psi_i(\phi(G_{\pi}^t))}{\lVert \psi_i(\phi(G_{\pi}^t)) \rVert}$
    \FOR{$(p, q) \in \mathcal{M}$ such that $s_{p, q} > \tau$}
      \STATE $\text{Set the class label of the merged object, }\hat{c}$
      \STATE $\hat{o}_{m} \gets (\hat{c}, \mathcal{A}^s_p \cup \mathcal{A}^t_q)$
      \STATE $\mathcal{O}^m \gets \{\hat{o}_{m}\} \cup \bigl(\mathcal{O}^m \setminus \{o^s_p, o^t_q\}\bigr)$
      \STATE $\text{Update  }\mathcal{E}^m: e_{m, *} \leftarrow e_{p,*}~\text{and}~e_{*, m} \leftarrow e_{*,q}$
    \ENDFOR
    \STATE $\mathcal{G} \gets \{G^m\} \cup (\mathcal{G} \setminus \{G^s, G^t\}) $
  \ENDWHILE
  \STATE $G_{\text{video}} \gets \text{extract}(\mathcal{G})$
  \STATE \textbf{return} $G_{\text{video}}$
\end{algorithmic}
\end{algorithm} 
The scene graph consolidation step combines all individual scene graphs derived from each segment into a unified graph that represents the overall visual content of the video.
We first describe our graph merging procedure and then introduce a subgraph extraction technique designed to support more focused and coherent video caption generation.

\subsubsection{Merging two scene graphs}
We first describe our scene graph merging technique.
Given two scene graphs, $G^s = (\mathcal{O}^s, \mathcal{E}^s)$ and $G^t = (\mathcal{O}^t, \mathcal{E}^t)$, constructed from captions corresponding to two different segments, we run the Hungarian algorithm to obtain an optimal matching between the two object sets, $\mathcal{O}^s$ and $\mathcal{O}^t$, which is formally expressed as
\begin{equation}
	\pi^* = \underset{\pi \in \Pi}{\arg\max} \sum_{i} \frac{ \psi_i(\phi(G^s))}{\| \psi_i(\phi(G^s)) \|} \cdot \frac{\psi_i(\phi(G_\pi^t)) }{\| \psi_i(\phi(G_\pi^t)) \|},
\end{equation}
where $\phi(\cdot)$ denotes a graph encoder, $\psi_i(\cdot)$ is a function to extract the $i^\text{th}$ object from an embedded graph, and $\pi \in \Pi$ indicates a permutation of objects in a graph.
Note that the object matching is based on their cosine similarity, where we introduce dummy objects  to deal with different numbers of objects for matching.

After computing all matching pairs using the Hungarian algorithm, we identify a set of valid matches $\mathcal{M}$ by selecting object pairs $(o_p^s, o_q^t)$ whose similarity score $s_{p,q}$ exceeds a predefined threshold $\tau$.
For each valid match $(p,q) \in \mathcal{M}$, the merged object $\hat{o}_{m} \in \hat{\mathcal{O}}$ is defined as
\begin{equation}
    \hat{o}_{m} = (\hat{c} , \mathcal{A}^s_p \cup \mathcal{A}^t_q) \in \hat{\mathcal{O}},
\end{equation}
where $\hat{c}$ denotes a class label of a merged object and $\hat{\mathcal{O}}$ represents the set of all merged objects obtained from valid matches.
Note that $\hat{c}$ may differ from the original class label of $o_p^s$ or $o_q^t$.
This procedure results in a new merged scene graph, $G^m=(\mathcal{O}^m,\mathcal{E}^m)$, which combines each valid pair of matched objects, creating a new object.

We perform graph merging by iteratively selecting and consolidating pairs of graphs based on their embedding similarity.
In each iteration, the two most similar graphs are merged into a single graph, which replaces the original pair in the set of graphs.
This process is repeated until only one unified scene graph remains.
The final scene graph provides a comprehensive representation of the entire video that preserves detailed information from individual segments.
Algorithm~\ref{alg:hierarchical_graph_merge} describes the detailed procedure of our graph consolidation strategy.
  
\subsubsection{Prioritized subgraph extraction}
\label{sec:subgraph}
When concise and focused video captions are desired, we apply subgraph extraction to retain only the most contextually relevant information.
During the graph merging process, we track each node's merge count as a measure of its significance within the consolidated graph. 
We then identify the top $k$ nodes with the highest merge counts and extract their corresponding subgraphs. 
This approach prioritizes objects that consistently appear across multiple frames, as they often represent key entities in the scene. 
By focusing on salient elements and filtering out irrelevant details, our method constructs a compact scene graph that enables more focused video captioning.

\section{Video Caption Generation}
\label{sec:videocaption}
Our ultimate goal is to generate captions from a consolidated scene graph. 
To this end, we develop a graph-to-text decoding model trained on a dataset of graph-text pairs. 
At inference time, the model takes the consolidated scene graph representing the entire video as input and generates a caption that describes the video as a whole.

\subsection{Graph-to-text model}
\label{sec:g2t_model}
Our graph-to-text model consists of a transformer-based graph encoder and a text decoder. 
The encoder processes the input scene graph to produce a graph embedding, which conditions the decoder to generate the final caption. 
To reflect the graph topology in our model, we design an attention mask in the graph encoder that restricts attention propagation to the edges defined in the scene graph.

To construct input tokens for the graph encoder, we convert the text values associated with each graph component, such as object classes $c_i$, attribute sets  $\mathcal{A}_i$, and edge labels $r_{i,j}$ (\eg, ``elderly'', ``woman'', ``cook in'', ``kitchen''), to sequences of embedding vectors.
Additionally, we append a learnable embedding token that attends to all other tokens, enabling the aggregation of global context and facilitating information flow across the entire graph, including between disconnected nodes.

\subsection{Training}
\label{sec:training}
We train the graph-to-text model on a large-scale collection of graph-text pairs. 
To construct this dataset, we curated approximately 2.5 million captions from diverse image captioning datasets, including MS-COCO~\cite{chen2015microsoft}, Flickr30k~\cite{young2014image}, TextCaps~\cite{sidorov2020textcaps}, Visual Genome~\cite{krishna2017visual}, and Visual Genome Paragraph Captions~\cite{krause2016paragraphs}, to cover a broad range of visual scene contexts.
To further enrich the dataset, we incorporated model-generated captions for videos in Kinetics-400~\cite{kay2017kinetics}, where LLaVA-NeXT-7B~\cite{liu2024llavanext} is applied to four uniformly sampled frames per video.
Each caption is then parsed into a scene graph using a textual scene graph parser, yielding a graph-text pair for training.

Using the graph-text pairs, we train the graph-to-text decoder with a next-token prediction objective, aiming to generate the ground-truth caption conditioned on the input scene graph, as formally defined below:
\begin{equation}
\mathcal{L}(\theta) = \sum_{i=1}^{N} \log P_{\theta}(t_i \mid t_{1:i-1}, G),
\end{equation}  
where $t_i$ represents the $i^\text{th}$ token in the source text, and $N$ denotes the total number of tokens.


\section{Experiment}
\label{sec:experiments}
This section presents the effectiveness of the proposed approach through performance evaluation and analysis on both video captioning and video paragraph captioning datasets.

\subsection{Experimental setup}
\label{sec:exp_setup}
We provide the detailed information about target tasks with their datasets and baselines.  
We also discuss a list of performance metrics used in our evaluation.

\subsubsection{Target tasks and baselines}
Our evaluation consists of two zero-shot tasks: (1) video captioning, using the standard test splits of MSR-VTT~\cite{xu2016msr-vtt} and MSVD~\cite{chen2011collecting}, and (2) video paragraph captioning, using the \textit{ae-val} set of ActivityNet Captions~\cite{krishna2017dense}, which contains longer videos with multiple events.

We primarily compare our method against LLM-based approaches. 
Specifically, we first establish an LLM summarization baseline, which directly summarizes the same set of segment-level captions used by our method. 
This baseline provides a direct comparison between the proposed scene graph consolidation and the simple aggregation of segment-level captions by LLMs.
We use the open-source Mistral-7B-Instruct-v0.3\footnote{\url{https://huggingface.co/mistralai/Mistral-7B-Instruct-v0.3}} for all datasets. 
For the ActivityNet Captions dataset, we additionally employ GPT-4o mini, a more powerful proprietary model.
Details of the prompt instructions used for the LLM summarization baselines are provided in Appendix~\ref{appendix_sec:prompt}.

We also compare our method against LLM-based video understanding methods, \eg, VidIL~\cite{wang2022language} and Video ChatCaptioner~\cite{chen2023video}, which utilize commercial LLMs along with textual representations derived from VLMs.
VidIL constructs rich input sequences by combining various textual cues such as objects, events and frame captions extracted from multiple image-based VLMs, and incorporates few-shot exemplars to guide the LLM in generating video captions.
Similarly, Video ChatCaptioner adopts an interactive question-answering framework between image VLM and LLMs.

Note that we primarily focus on LLM-based approaches, as other approaches typically require supervised fine-tuning, making direct zero-shot comparisons infeasible. 
Additional comparisons with broader zero-shot video captioning approaches---for example, test-time optimization, inference optimization, and text-only training methods---on MSR-VTT are included in the supplementary document.

\subsubsection{Evaluation metrics}
Following standard performance evaluation protocols in video captioning, our experiments adopt $n$-gram-based metrics, including BLEU-4 (B@4)~\cite{papineni2002bleu}, METEOR~\cite{banerjee2005meteor}, and CIDEr~\cite{vedantam2015cider}, which measure the overlap between generated and reference captions. 
Since these $n$-gram-based metrics are limited in capturing semantic details and contextual accuracy beyond literal phrase matching, 
we additionally employ BERTScore~\cite{zhang2019bertscore}, an embedding-based evaluation metric widely used in natural language processing tasks such as machine translation and summarization.
BERTScore measures token-level cosine similarities between generated and reference captions, capturing semantic similarity beyond $n$-gram matches as follows: 
\begin{align}
	P_{\text{BERT}} &= \frac{1}{|\hat{\mathcal{Z}}|} \sum_{\hat{z}_j \in \hat{\mathcal{Z}}} \max_{z_i \in \mathcal{Z}} z_i^{\top} \hat{z}_j, \\
	R_{\text{BERT}} &= \frac{1}{|\mathcal{Z}|} \sum_{z_i \in \mathcal{Z}} \max_{\hat{z}_j \in \hat{\mathcal{Z}}} z_i^{\top} \hat{z}_j, \\
	F_{\text{BERT}} &= \frac{2 \cdot P_{\text{BERT}} \cdot R_{\text{BERT}}}{P_{\text{BERT}} + R_{\text{BERT}}},
\end{align}
where $\mathcal{Z} \equiv \{ z_1, z_2, \dots\}$ and $\hat{\mathcal{Z}} \equiv \{ \hat{z}_1, \hat{z}_2, \dots\}$ represent the sets of token embeddings in the reference and generated captions, respectively. 
%


\begin{table*}[!t]
	\centering
	\renewcommand{\arraystretch}{1.05}
	\setlength{\tabcolsep}{9pt}
	\caption{
	Zero-shot video captioning results on the MSR-VTT~\cite{xu2016msr-vtt} and MSVD~\cite{chen2011collecting} test sets, comparing our method (SGVC) with LLM-based video understanding methods.
	$\dagger$~indicates that the method utilizes reference captions from the target dataset to construct few-shot exemplar prompts. 
	Bold numbers indicate the highest scores among methods not using reference captions.
	}
	\vspace{2mm}
	\label{table:vc_comparison}
	\scalebox{0.85}{
\begin{tabular}{@{}llccccccc@{}}
		\toprule
		Dataset & Method & Backbone VLM & B@4 & METEOR & CIDEr & $P_{\text{BERT}}$ & $R_{\text{BERT}}$ & $F_{\text{BERT}}$ \\
		\midrule
		\multirow{5}{*}{\makecell{MSR-VTT}} &VidIL~\cite{wang2022language} &\multirow{2}{*}{\makecell{BLIP+CLIP}} & 3.2 & 14.8 & 3.1 & 0.134 & 0.354 & 0.225 \\	
		 & VidIL$^\dagger$~\cite{wang2022language}  & & 13.6 & 20.0 & 20.2 & 0.461 & 0.552 & 0.490 \\	
		\cdashline{2-9}[0.5pt/1.0pt]
		& Video ChatCaptioner~\cite{chen2023video} & BLIP2 & 13.2 & 22.0 & 16.5 & 0.396 & 0.510 & 0.436 \\	
		\cdashline{2-9}[0.5pt/1.0pt]
		& \multirow{2}{*}{\makecell{\textbf{SGVC (Ours)}}} & BLIP & 17.7 & 22.5 & 24.0 & \textbf{0.476} & 0.539 & \textbf{0.490} \\
		& & BLIP2 & \textbf{18.4} & \textbf{23.1} & \textbf{26.1} & 0.467 & \textbf{0.542} & 0.487 \\
		\midrule
		\multirow{5}{*}{\makecell{MSVD}} & VidIL~\cite{wang2022language} &\multirow{2}{*}{\makecell{BLIP+CLIP}} & 2.5 & 16.5 & 2.3 & 0.124 & 0.404 & 0.238 \\	
		 & VidIL$^\dagger$~\cite{wang2022language} & & 30.7 & 32.0 & 60.3 & 0.656 & 0.726 & 0.674\\	
		\cdashline{2-9}[0.5pt/1.0pt]
		& Video ChatCaptioner~\cite{chen2023video} & BLIP2 & 22.7 & 31.8 & 35.8 & 0.496 & 0.651 & 0.550 \\	
		\cdashline{2-9}[0.5pt/1.0pt]
		& \multirow{2}{*}{\makecell{\textbf{SGVC (Ours)}}} & BLIP & 22.6 & 30.2 & 50.2 & \textbf{0.575} & 0.646 & 0.589 \\
		& & BLIP2 & \textbf{25.3} & \textbf{32.0} & \textbf{53.3} & 0.571 & \textbf{0.669} & \textbf{0.597}\\
		\bottomrule
		\end{tabular}
		}
		\vspace{-2mm}
\end{table*}

\begin{table*}[!t]
	\centering
	\renewcommand{\arraystretch}{1.05}
	\setlength{\tabcolsep}{11.5pt}
	\caption{
	Zero-shot video captioning results on the MSR-VTT~\cite{xu2016msr-vtt} and MSVD~\cite{chen2011collecting} test sets, comparing SGVC with the LLM summarization baseline.
	Bold numbers indicate the highest scores.
	}
	\vspace{2mm}
	\label{table:vc_llm_summ}
	\scalebox{0.85}{
\begin{tabular}{@{}llccccccc@{}}
		\toprule
		Dataset & Method & Backbone VLM & B@4 & METEOR & CIDEr & $P_{\text{BERT}}$ & $R_{\text{BERT}}$ & $F_{\text{BERT}}$ \\
		\midrule
		 \multirow{4}{*}{\makecell{MSR-VTT}} & \multirow{2}{*}{\makecell{Summarization w/ Mistral-7B}} & BLIP & 9.6 & 21.6 & 10.8 & 0.313 & 0.516 & 0.395 \\
		& & BLIP2 & 11.5 & \textbf{23.1} & 15.4 & 0.308 & 0.528 & 0.397 \\ 
		\cdashline{2-9}[0.5pt/1.0pt]
		& \multirow{2}{*}{\makecell{\textbf{SGVC (Ours)}}} & BLIP & 17.7 & 22.5 & 24.0 & \textbf{0.476} & 0.539 & \textbf{0.490} \\
		& & BLIP2 & \textbf{18.4} & \textbf{23.1} & \textbf{26.1} & 0.467 & \textbf{0.542} & 0.487 \\
		\midrule		
		 \multirow{4}{*}{\makecell{MSVD}}  & \multirow{2}{*}{\makecell{Summarization w/ Mistral-7B}} & BLIP & 15.2 & 28.3 & 30.3 & 0.477 & 0.623 & 0.527 \\
		& & BLIP2 & 22.5 & 31.9 & 41.6 & 0.500 & 0.664 & 0.558 \\
		\cdashline{2-9}[0.5pt/1.0pt]
		& \multirow{2}{*}{\makecell{\textbf{SGVC (Ours)}}} & BLIP & 22.6 & 30.2 & 50.2 & \textbf{0.575} & 0.646 & 0.589 \\
		& & BLIP2 & \textbf{25.3} & \textbf{32.0} & \textbf{53.3} & 0.571 & \textbf{0.669} & \textbf{0.597}\\
		\bottomrule
		\end{tabular}
		}
		\vspace{-2mm}
\end{table*}

\subsection{Implementation details}
Our graph-to-text model consists of a graph encoder and a text decoder, with a total of 235M parameters.
The BERT-base model~\cite{devlin2018bert} is employed for our encoder, with attention masking as described in Section~\ref{sec:g2t_model}, and only the decoder part of T5-base~\cite{raffel2020exploring} is used as our text decoder.

The graph-to-text model is trained on graph-text pairs constructed in Section~\ref{sec:training} for $1K$ iterations with a batch size of 512.
We employ the AdamW~\cite{loshchilov2017decoupled} optimizer with a weight decay of 0.05, an initial learning rate of 0.0001, and linear warm-up for the first 1\% of training steps. 
For video paragraph captioning, the model is further fine-tuned for 400 iterations on the subset of the constructed graph-text pairs obtained from the Visual Genome paragraph captioning dataset~\cite{krause2016paragraphs}.

Segment-level captions are generated using off-the-shelf VLMs.
To demonstrate the flexibility of our approach, we employed both image-centric VLMs, including BLIP~\cite{li2022blip} and BLIP2~\cite{li2023blip}, and video-centric VLM, InternVL2.5~\cite{chen2024expanding}. 
For MSR-VTT and MSVD, we uniformly sample six frames per video to generate captions using image-centric models.
For ActivityNet Captions, we select twelve frames per video when using image-centric VLMs, while extracting twelve video clips for the video-centric model.

For generating the final video caption, we apply a beam search with five beams, a maximum sequence length of 32 and a length penalty of 0.6.
For video captioning on MSR-VTT, we apply prioritized subgraph extraction with $k=1$ to emphasize salient visual information.
Video paragraph caption, which requires more detailed descriptions, is generated using a beam search with three beams, a maximum sequence length of 400, and a repetition penalty of 3.0.


\begin{table*}[!t]
	\centering
	\renewcommand{\arraystretch}{1.05}
	\setlength{\tabcolsep}{13pt}
	\caption{Zero-shot video paragraph captioning results on the ActivityNet Captions~\cite{krishna2017dense} \text{\it{ae-val}} set, comparing our method (SGVC) with LLM-based video understanding methods. 
	$\dagger$~indicates that the method utilizes reference captions from the target dataset to construct few-shot exemplar prompts. 
 	Bold numbers indicate the highest scores among methods not using reference captions.
	}
	\vspace{2mm}
	\label{table:vpc_comp}
	\scalebox{0.85}{
	\begin{tabular}{@{}lccccccc@{}}
		\toprule
		Method & Backbone VLM & B@4 & METEOR & CIDEr & $P_{\text{BERT}}$ & $R_{\text{BERT}}$ & $F_{\text{BERT}}$ \\
		\midrule
		 VidIL~\cite{wang2022language} & \multirow{2}{*}{\makecell{BLIP+CLIP}} & 1.0 & 5.8 & 4.6 & 0.122 & 0.135 & 0.125 \\	
		 VidIL$^\dagger$~\cite{wang2022language} & & 2.9 & 7.6 & 3.3 & 0.414 & 0.243 & 0.323\\	
       		 \cdashline{1-8}[0.5pt/1.0pt]
		Video ChatCaptioner~\cite{chen2023video} & BLIP2 & 2.4 & 8.9 & 1.6 & 0.207 & 0.202 & 0.200 \\		
		\midrule		
		\multirow{2}{*}{\makecell{\textbf{SGVC (Ours)}}} & BLIP & 6.7 & 11.6 & 16.6 & \textbf{0.367} & 0.285 & 0.322 \\
		 & BLIP2 & \textbf{7.4} & \textbf{12.4} & \textbf{20.9} & \textbf{0.367} & \textbf{0.304} & \textbf{0.331} \\
		\bottomrule
		\end{tabular}
		}
		\vspace{-2mm}
\end{table*}

\begin{table*}[!t]
	\centering
	\renewcommand{\arraystretch}{1.05}
	\setlength{\tabcolsep}{15.5pt}
	\caption{Zero-shot video paragraph captioning results on the ActivityNet Captions~\cite{krishna2017dense} \text{\it{ae-val}} set, comparing SGVC with the LLM summarization baselines. 
	Bold numbers indicate the highest scores.
	}
	\vspace{2mm}
	\label{table:vpc2_llm}
	\scalebox{0.85}{
	\begin{tabular}{@{}lccccccc@{}}
		\toprule
		Method & Backbone VLM & B@4 & METEOR & CIDEr & $P_{\text{BERT}}$ & $R_{\text{BERT}}$ & $F_{\text{BERT}}$ \\
		\midrule
		\multirow{3}{*}{\makecell{Summarization w/ Mistral-7B}} & BLIP & 3.4 & 9.4 & 7.5 & 0.292 & 0.268 & 0.276  \\
		 & BLIP2 & 4.1 & 10.4 & 9.6 & 0.307 & 0.293 & 0.295 \\
		 & InternVL2.5 & 4.5 & 10.8 & 11.6 & 0.333 & 0.318 & 0.319 \\
		\midrule
		\multirow{3}{*}{\makecell{Summarization w/ GPT-4o mini}} & BLIP & 4.6 & 10.2 & 10.3 & 0.325 & 0.284 & 0.300  \\
		 & BLIP2 & 5.0 & 10.6 & 12.1 & 0.343 & 0.301 & 0.317 \\
		 & InternVL2.5 & 5.8 & 11.4 & 15.3 & 0.352 & \textbf{0.332} & 0.336 \\
		\midrule
		 \multirow{3}{*}{\makecell{\textbf{SGVC (Ours)}}} & BLIP & 6.7 & 11.6 & 16.6 & \textbf{0.367} & 0.285 & 0.322 \\
		 & BLIP2 & 7.4 & 12.4 & 20.9 & \textbf{0.367} & 0.304 & 0.331 \\
		 & InternVL2.5 & \textbf{8.0} & \textbf{13.2} & \textbf{24.1} & 0.359 & 0.326 & \textbf{0.338} \\
		\bottomrule
		\end{tabular}
		}
		\vspace{-2mm}
\end{table*}


\begin{figure*}[!t]
    \centering
        \begin{tabular}{@{}cc@{}}
                \includegraphics[width=0.485\linewidth]{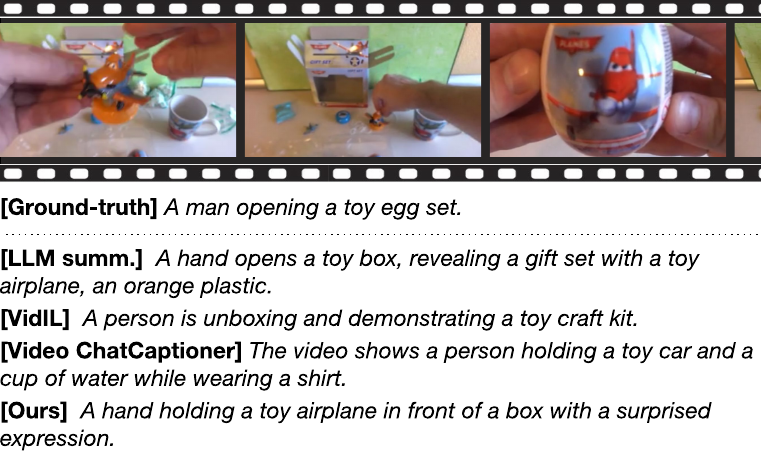} &
                \includegraphics[width=0.485\linewidth]{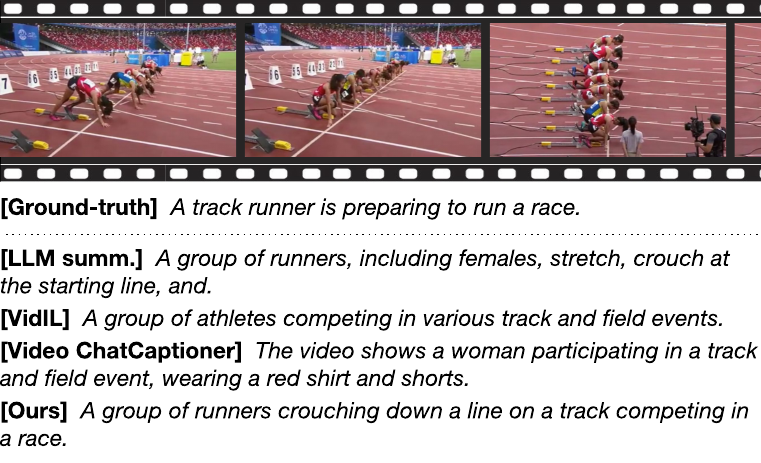} \\
		\vspace{-3mm} \\
                \includegraphics[width=0.485\linewidth]{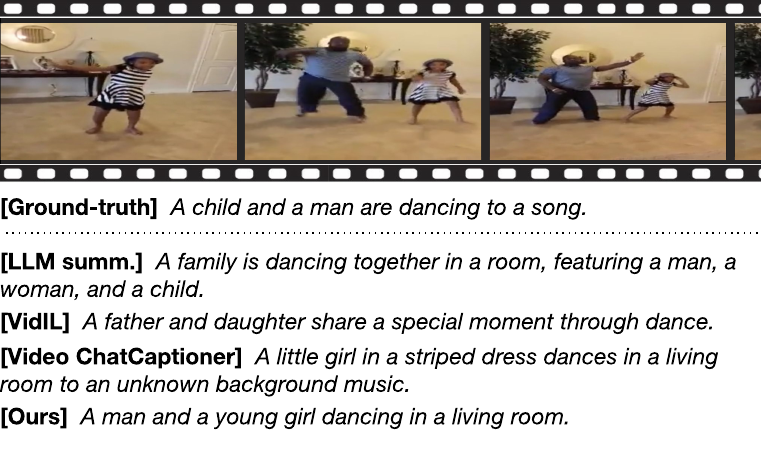} &
                \includegraphics[width=0.485\linewidth]{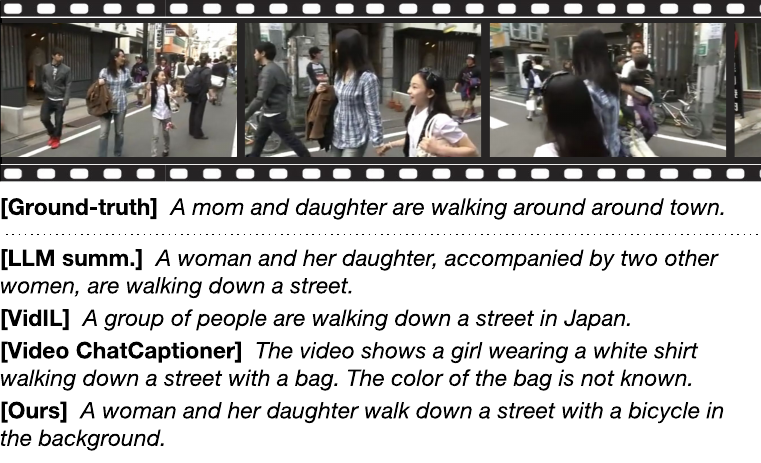} \\
                \vspace{-3mm} \\
                \includegraphics[width=0.485\linewidth]{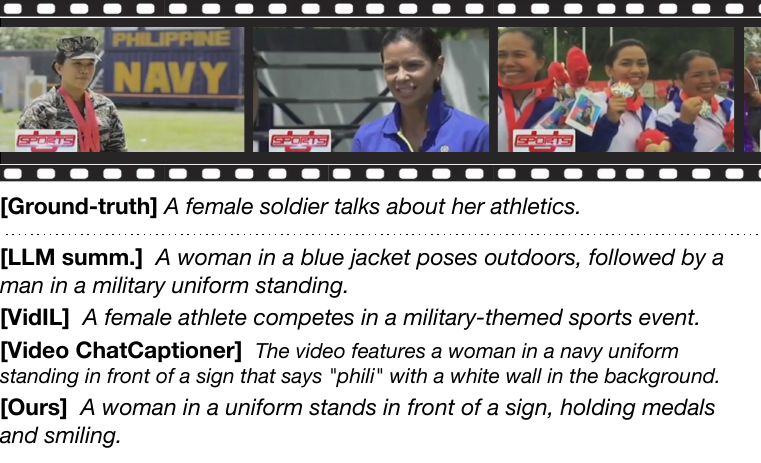} &
                \includegraphics[width=0.485\linewidth]{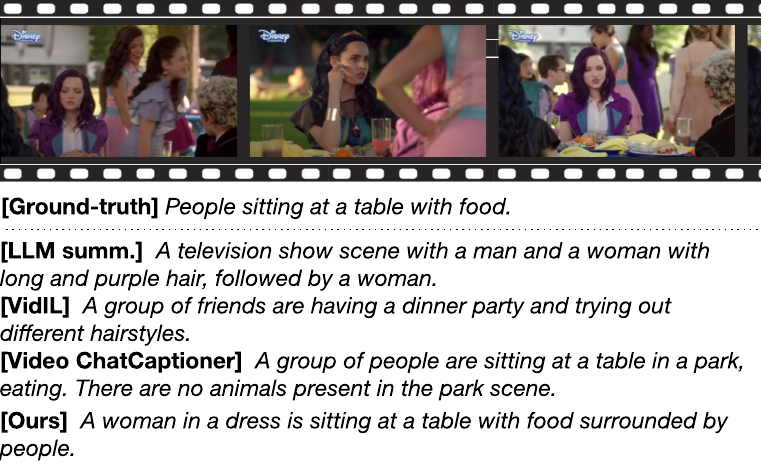} \\
        \end{tabular}
    \vspace{-2mm}
    \caption{
    Example of zero-shot video captioning results on the MSR-VTT test set. 
    We compare our results with LLM-based methods, listed from top to bottom as 1) LLM summarization using Mistral-7B, 2) VidIL, 3) Video ChatCaptioner, and 4) SGVC (Ours).
    }
    \vspace{-5mm}
    \label{fig:qual_vc}
\end{figure*}

\subsection{Main results}
We present quantitative results for zero-shot video captioning on the MSR-VTT and MSVD datasets in Tables~\ref{table:vc_comparison} and \ref{table:vc_llm_summ}, and for zero-shot video paragraph captioning on the ActivityNet Captions \textit{ae-val} set in Tables~\ref{table:vpc_comp} and \ref{table:vpc2_llm}.

\subsubsection{Zero-shot video captioning}

Table~\ref{table:vc_comparison} compares the proposed method, SGVC, with existing LLM-based video understanding approaches, VidIL and Video ChatCaptioner.
SGVC consistently achieves strong zero-shot performance across most metrics on both the MSR-VTT and MSVD datasets, outperforming the existing methods.  
VidIL, although it leverages diverse textual cues from multiple sources, shows limited performance in the zero-shot setting. 
Notably, SGVC performs competitively even against VidIL's few-shot setting, which heavily depends on dataset-specific exemplars.  
Video ChatCaptioner, which aggregates information through multi-turn question answering between an LLM and BLIP2, often suffers from hallucinations or overemphasis on irrelevant details, leading to failures in capturing the core content of the video (\eg, ``There are no animals present in the park scene.").

Table~\ref{table:vc_llm_summ} provides a controlled comparison between SGVC and an LLM-based summarization method, clearly highlighting the effectiveness of our scene graph consolidation approach.
Both methods start from an identical set of segment-level captions and this experiments isolates the impact of the graph consolidation.
Although LLM summarization produces fluent and expressive captions, it sometimes overlooks details of objects and events within a scene.
In contrast, SGVC explicitly integrates segment-level scene graphs into a unified representation, which is helpful for preserving object identities and relationships consistently throughout the video.


\begin{table*}[!t]
	\centering
	\renewcommand{\arraystretch}{1.05}
	\setlength{\tabcolsep}{7pt}
	\caption{Comparison of computational costs between SGVC and LLM-based methods on the MSR-VTT test set.}
	\vspace{2mm}
	\label{table:efficiency}
	\scalebox{0.85}{
		\begin{tabular}{@{}lcccccccc@{}}
		\toprule
		Method & VLM Backbone & Params. (B) & GPU (GB) & Time (s) & CIDEr & Using reference & Using GPT API \\
		\midrule
		VidIL & BLIP+CLIP & 0.67 & 3.57 & 1.32 & 20.2 & \checkmark & \checkmark \\
		Video ChatCaptioner & BLIP2 & 3.75 & 14.53 & 3.65 & 16.5 & - & \checkmark \\
		\midrule
		\multirow{2}{*}{\makecell{Summarization w/ Mistral-7B}} & BLIP & 7.50 & 14.50 & 1.27 & 10.8 & - & - \\
		& BLIP2 & 11.00 & 28.20 & 1.51 & 15.4 & -- & -- \\
		\midrule
		\multirow{2}{*}{\makecell{\textbf{SGVC (Ours)}}} & BLIP & 0.74 & 5.07 & 1.14 & 24.0 & - &  - \\
		 & BLIP2 & 4.24 & 18.40 & 1.37 & 26.1 & -- & -- \\
		\bottomrule
	\end{tabular}
	}
	\vspace{-2mm}
\end{table*}

\subsubsection{Zero-shot video paragraph captioning}
Table~\ref{table:vpc_comp} presents a comparison between SGVC and other LLM-based video understanding methods for zero-shot video paragraph captioning on the ActivityNet Captions \textit{ae-val} set.
Consistent with the results observed in zero-shot video captioning in Table~\ref{table:vc_comparison}, SGVC clearly outperforms competing methods. 
The performance gap is even more pronounced in the paragraph captioning task, where effectively modeling long-range context and maintaining coherence across multiple events is essential.

Table~\ref{table:vpc2_llm} compares SGVC with LLM summarization techniques, using both Mistral-7B and a stronger commercial model, GPT-4o mini. 
While GPT-4o mini offers significant performance gains over Mistral-7B, it still falls short of SGVC, highlighting the effectiveness of our graph consolidation approach.
Furthermore, replacing the backbone captioner with InternVL2.5 further improves SGVC's performance, benefiting from its video-centric design and strong temporal modeling capabilities, despite having significantly fewer parameters than BLIP2 (938M vs. 3.74B).
These results clearly demonstrate SGVC's flexibility and plug-and-play compatibility  with a wide range of vision-language model architectures.

\subsection{Analysis}
\paragraph{Efficiency}
Table~\ref{table:efficiency} presents a detailed comparison of computational costs, in terms of average per-video inference time and peak GPU memory usage on a single NVIDIA A6000 GPU, along with captioning performance (CIDEr) on the MSR-VTT test set. 
SGVC consistently outperforms LLM-based summarization approaches across all computational measures, regardless of the underlying backbones.
Moreover, our scene graph merging algorithm, which currently runs on the CPU, could be further accelerated by GPU implementation.
VidIL and Video ChatCaptioner exhibit slower inference times and lower captioning accuracy. 
While they consume less GPU memory, their dependence on GPT API calls introduces additional latency.


\begin{table}[t]
	\centering
	\renewcommand{\arraystretch}{1.05}
	\setlength{\tabcolsep}{12pt}
	\caption{Analysis on the hyperparameter $k$ in the prioritized subgraph extraction, on the MSR-VTT test set. 
	}
	\vspace{2mm}
	\label{table:top-k}
	\scalebox{0.85}{
	\begin{tabular}{@{}cccccc@{}}
	\toprule
	$k$ & METEOR & CIDEr & $P_{\text{BERT}}$ & $R_{\text{BERT}}$ & $F_{\text{BERT}}$ \\
	\midrule
	1 & 23.1 & \textbf{26.1} & \textbf{0.467} & 0.542 & \textbf{0.487} \\
	3 & \textbf{23.8} & 24.9 & 0.454 & \textbf{0.554} & 0.486 \\	
	\bottomrule
	\end{tabular}
	}
	\vspace{-5mm}
\end{table}
\paragraph{Impact of hyperparameters}
We analyze the effect of the hyperparameter $k$, which controls the size of the extracted subgraph, as described in Section~\ref{sec:subgraph}.
%
\begin{table}[t]
	\centering
	\renewcommand{\arraystretch}{1.05}
	\setlength{\tabcolsep}{12pt}
	\caption{Analysis on the threshold $\tau$ used in graph consolidation, on the MSVD test set. 
	}
	\vspace{2.75mm}
	\label{table:tau}
	\scalebox{0.85}{
	\begin{tabular}{@{}ccc|ccc@{}}
	\toprule
	$\tau$ & CIDEr & $F_{\text{BERT}}$ & $\tau$ & CIDEr & $F_{\text{BERT}}$ \\
	\midrule
	0.95 & 50.0 & 0.589 & 0.85 & 49.9 & 0.589\\
	0.90 & 50.2 & 0.589 & 0.80 & 49.9 & 0.589 \\
	\bottomrule
	\end{tabular}
	}
	\vspace{-5.5mm}
\end{table}
As shown in Table~\ref{table:top-k}, lower $k$ values result in more concise subgraphs that emphasize salient objects, leading to improvements in precision-oriented metrics, such as CIDEr and $P_{\text{BERT}}$.

\begin{figure*}[t]
    \centering
        \begin{tabular}{@{}c@{}}
                \includegraphics[width=1\linewidth]{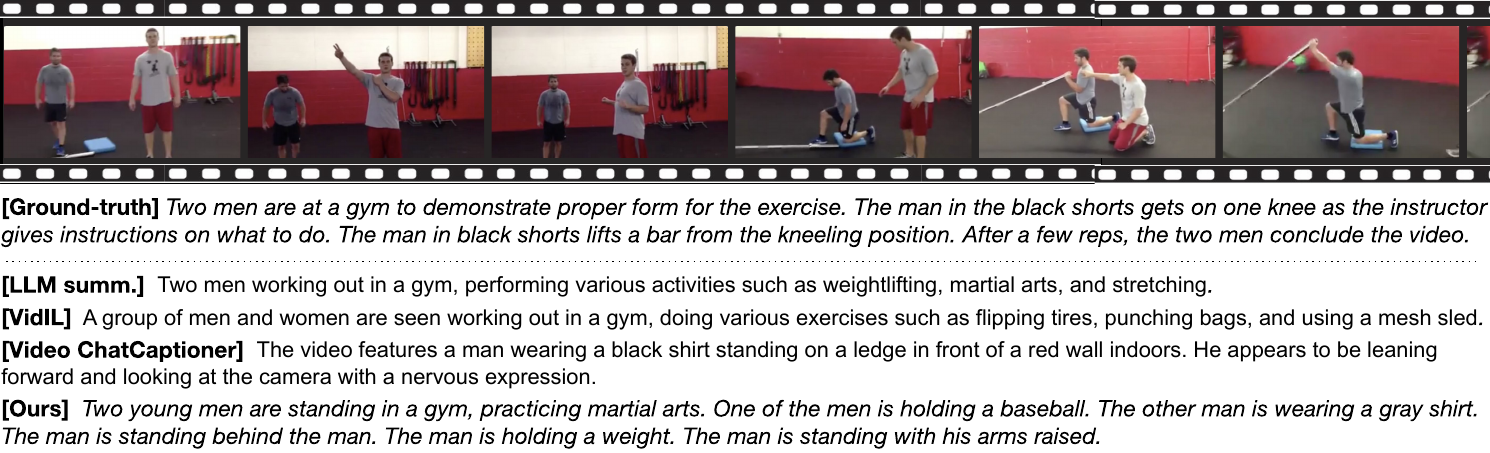} \\ 
                \vspace{-1mm} \\
                \includegraphics[width=1\linewidth]{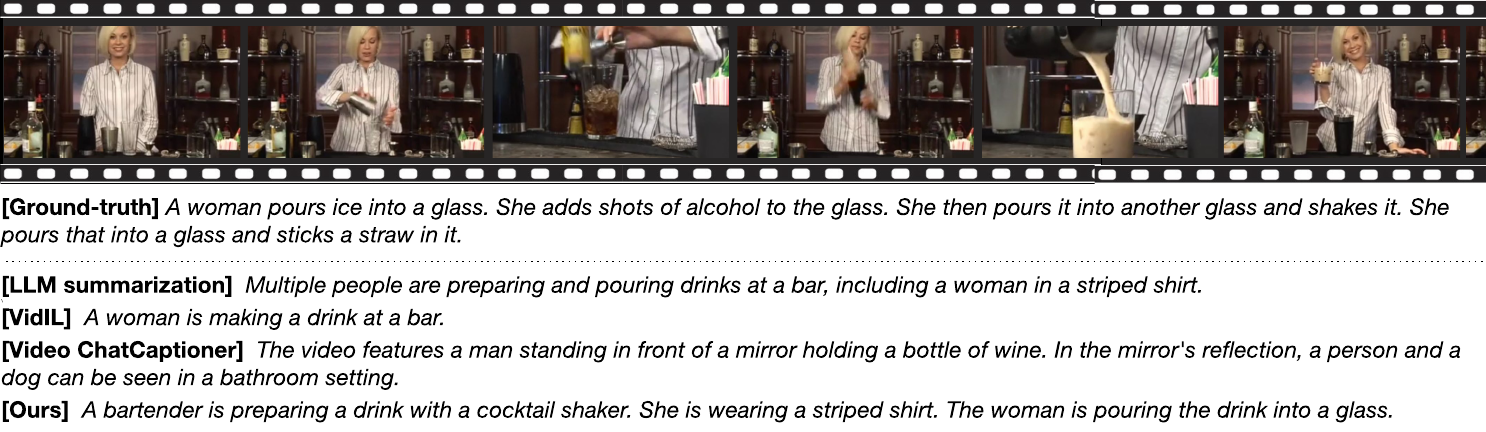} \\
            \end{tabular}
    \vspace{-3mm}
    \caption{Example of zero-shot video paragraph captioning results on the \text{\it{ae-val}} set of the ActivityNet captions dataset.
    We compare our results with LLM-based methods, listed from top to bottom as 1) LLM summarization using Mistral-7B, 2) VidIL, 3) Video ChatCaptioner, and 4) SGVC (Ours).
    }
     \vspace{-4mm}
    \label{fig:qual_vpc}
\end{figure*}

In contrast, higher $k$ values yield richer subgraphs that capture broader contextual information, thereby improving recall-oriented metrics, METEOR and $R_{\text{BERT}}$.

We also conducted evaluation by varying the cosine similarity threshold $\tau$, as reported in Table~\ref{table:tau}.
The results demonstrate stable performance within the range $\tau \in [0.80, 0.95]$, and we set $\tau=0.9$ for all experiments.

\paragraph{Qualitative results}
Figures~\ref{fig:qual_vc} and~\ref{fig:qual_vpc} present qualitative examples of zero-shot video captioning on the MSR-VTT test set and video paragraph captioning on ActivityNet Captions \textit{ae-val} set, respectively.
Our method generates detailed and contextually rich captions that accurately capture events, objects, and relationships across frames.
While LLM summarization and Video ChatCaptioner produce fluent sentences, they occasionally introduce hallucinated content, such as objects or attributes that are not actually present in the video.


\section{Conclusion}
\label{sec:conclusion}

We introduced a novel framework for fine-grained captioning of long videos by consolidating information across multiple temporal segments.
Our approach merges scene graphs extracted from segment-level captions to generate comprehensive and coherent video descriptions.
This framework provides a computationally efficient and training-free alternative to existing methods. 
In contrast to LLM-based approaches, our method significantly reduces computational demands by leveraging a lightweight graph-to-text model with substantially fewer parameters. 
Extensive experiments on both video captioning and video paragraph captioning tasks validate the effectiveness of our method.
These results highlight the potential of graph-based consolidation as a foundation for future advances in long video captioning.

\section*{Acknowledgements}
We thank Do Young Eun at North Carolina State University for the valuable discussions. 
This work was supported in part by National Research Foundation of Korea (NRF) grant [RS-2022-NR070855, Trustworthy Artificial Intelligence], Institute of Information \& communications Technology Planning \& Evaluation (IITP) grants [RS2022-II220959 (No.2022-0-00959), (Part 2) Few-Shot Learning of Causal Inference in Vision and Language for Decision Making; No.RS-2021-II212068, AI Innovation Hub (AI Institute, Seoul National University); No.RS-2021-II211343, Artificial Intelligence Graduate School Program (Seoul National University)] funded by the Korea government (MSIT), and by Brain Pool program funded by the Ministry of Science and ICT through the National Research Foundation of Korea (No. RS-2024-00408610).

\section*{Impact Statement}
The broader impact of this research lies in enabling effective captioning of long videos by leveraging existing vision-language models without any additional fine-tuning on large-scale annotated video datasets.
While there is potential for societal impacts arising from this technology, we have not identified any significant negative consequences directly associated with our approach.

\bibliography{main}
\bibliographystyle{icml2025}

\clearpage
\appendix

\section{Additional Experiment}

\begin{table*}[!t]
	\centering
	\vspace{-5mm}
	\renewcommand{\arraystretch}{1.1}
	\setlength{\tabcolsep}{9pt}
	\caption{Zero-shot video captioning results on the MSR-VTT test set~\cite{xu2016msr-vtt}. 
	\checkmark~indicates that the method utilizes reference captions from the MSR-VTT dataset. 
	* indicates methods were adapted to zero-shot video captioning by Tewel~\etal~\cite{Tewel_2023_BMVC}.
	Bold numbers indicate the highest scores among methods not using reference captions.
	}
	\vspace{2mm}
	\label{table:vc_supple_msrvtt}
	\scalebox{0.8}{
		\begin{tabular}{@{}lcccccccc@{}}
		\toprule
		Method & Backbone VLM & Using reference & B@4 & METEOR & CIDEr & $P_{\text{BERT}}$ & $R_{\text{BERT}}$ & $F_{\text{BERT}}$ \\
		\midrule
		\rowcolor{gray!15}\multicolumn{9}{c}{\textbf{\textit{Consolidation-based approaches}}} \\
		\vspace{-3mm} \\ 
       		 \multirow{2}{*}{\makecell{VidIL~\cite{wang2022language}}} &\multirow{2}{*}{\makecell{BLIP+CLIP}} & & 3.2 & 14.8 & 3.1 & 0.134 & 0.354 & 0.225 \\	
		 & & \checkmark & 13.6 & 20.0 & 20.2 & 0.461 & 0.552 & 0.490 \\	
		 \cdashline{1-9}[0.5pt/1.0pt]
		Video ChatCaptioner~\cite{chen2023video} & BLIP2 & & 13.2 & 22.0 & 16.5 & 0.396 & 0.510 & 0.436 \\	
		\midrule
		\multirow{3}{*}{\makecell{Summ. w/ Mistral-7B}} & BLIP & & 9.6 & 21.6 & 10.8 & 0.313 & 0.516 & 0.395 \\
		& BLIP2 & & 11.5 & 23.1 & 15.4 & 0.308 & 0.528 & 0.397 \\
		& LLAVA-Next-7B & & 15.3 & \textbf{23.8} & 19.5 & 0.338 & 0.535 & 0.414 \\
		\midrule
		\multirow{3}{*}{\makecell{\textbf{SGVC (Ours)}}} & BLIP & & 17.7 & 22.5 & 24.0 & \textbf{0.476} & 0.539 & 0.490 \\
		& BLIP2 & & \textbf{18.4} & 23.1 & \textbf{26.1} & 0.467 & 0.542 & 0.487 \\
		& LLAVA-Next-7B & & 17.1 & 23.0 & 24.0 & 0.455 & \textbf{0.547} & \textbf{0.497} \\	
		\midrule
		\rowcolor{gray!15}\multicolumn{9}{c}{\textbf{\textit{Other zero-shot video captioning approaches}}} \\
		\vspace{-3mm} \\ 
        		MAGIC*~\cite{su2022language} & CLIP & & 5.5 & 13.3 & 7.4 & - & - & - \\
		\midrule
		ZeroCap*~\cite{tewel2021zero} & \multirow{2}{*}{\makecell{CLIP}} & & 2.3 & 12.9 & 5.8 & - & - & - \\
        		Tewel et al.~\cite{Tewel_2023_BMVC} & & & 3.0 & 14.6 & 11.3 & 0.280 & 0.391 & 0.319 \\
        		\midrule
        		\vspace{-0.5cm} \\
	        Decap-BookCorpus~\cite{lidecap} & \multirow{3}{*}{\makecell{CLIP}} & & 6.0 & 12.7 & 12.3 & - & - & - \\
	        Decap-COCO~\cite{lidecap} & & & 14.7 & 20.4 & 18.6 & 0.429 & 0.537 & 0.465 \\
	        Decap-MSRVTT~\cite{lidecap} & & \checkmark & 23.1 & 23.6 & 34.8 & - & - & - \\
	        \cdashline{1-9}[0.5pt/1.0pt]
	        C$^\text{3}$~\cite{zhang2024connect} & ImageBind & \checkmark & 25.3 & 23.4 & 27.8 & 0.518 & 0.550 & 0.519\\
		\bottomrule
		\end{tabular}
		}
\end{table*}
We provide an extended comparison against a broader set of zero-shot video captioning methods on MSR-VTT test set in Table~\ref{table:vc_supple_msrvtt}.

We compared our approach with several existing approaches, including:
1) test-time optimization via gradient manipulation with CLIP embeddings, \eg, ZeroCap~\cite{tewel2021zero} and Tewel~\etal~\cite{Tewel_2023_BMVC}, 2) optimization of inference procedure in the decoder using the CLIP image-text similarity, \eg, MAGIC~\cite{su2022language}, and 3) text-only training methods, \eg, DeCap~\cite{lidecap} and C$^{3}$~\cite{zhang2024connect}, which are trained solely on text corpora, 4) LLM-based video understanding methods, \eg, VidIL~\cite{wang2022language} and Video ChatCaptioner~\cite{chen2023video}, which utilize proprietary, commercially available LLMs along with textual representations derived from various image-language models, and 5) LLM summarization, which takes the same set of segment-level captions as our method and generates video captions using a pretrained LLM, Mistral-7B-Instruct-v0.3 by text summarization.

Note that DeCap-MSRVTT, C$^{3}$, and VidIL all utilize annotations from the training dataset but differ in how these annotations are employed. 
Specifically, DeCap-MSRVTT and C$^{3}$ use text annotations from the MSR-VTT training set to train their text decoders.
In contrast, VidIL constructs few-shot exemplars to serve as prompts, enabling LLM\footnote{In all our experiments, we use GPT-3.5-turbo-instruct since text-davinci-002 has been deprecated.} to perform video captioning through in-context learning.

This comprehensive comparison demonstrates that our explicit scene-graph-based modeling achieves superior performance over existing zero-shot video captioning methods across all evaluation metrics.

\section{Prompt Instructions} 
We provide prompt instructions for segment-level caption generation and LLM summarization of these captions, illustrated here using an image-centric VLM for video captioning.
\label{appendix_sec:prompt}
\subsection{Segment caption generation} 
Table~\ref{appendix_tab:imageVLM_fc_prompt} lists the instructional prompts, generated using GPT-4, which guide VLM in generating the segment-level captions.
These prompts are designed to ensure captions remain grounded in the visible content of the image, thereby avoiding factual errors or hallucinated details not supported by the image.
A prompt was randomly selected for each segment, allowing captions to reflect diverse aspects of a video.
\begin{table*}[t]
	\centering
	\vspace{-2mm}
	\small
	\caption{The list of instructional prompts for segment-level caption generation using an image-centric VLM. 
	}
	\vspace{3mm}
	\label{appendix_tab:imageVLM_fc_prompt}
	\scalebox{0.95}{
		\fbox{
		    \begin{minipage}{0.8\linewidth}
		    {\linespread{1.2}\selectfont
		    \textbullet\ “Please describe what is happening in the image using one simple sentence. Focus only on what is visible.'' \par
   		    \textbullet\ “Now, provide a single sentence caption that describes only what is explicitly shown in the image” \par
   		    \textbullet\ “In one sentence, describe what you see in the image without adding any extra details.” \par
		    \textbullet\ “Provide a concise one-sentence description of the image, focusing on only the visible elements.” \par
  		    \textbullet\ “Please give a one-sentence caption that includes only what is clearly shown in the image.” \par
   		    \textbullet\ “Describe what is happening in the image in one simple sentence, without any added information.” \par
  		    \textbullet\ “Please generate a single sentence caption that describes only what can be seen in the image.” \par
  		    \textbullet\ “Provide a one-sentence description of the image, focusing solely on what is shown.” \par
		    \textbullet\ “Now, give a brief, one-sentence caption based strictly on the visible content in the image.” \par
  		    \textbullet\ “In a single sentence, describe what the image shows, without including anything extra.” \par
		    }
		    \end{minipage}
		}
	}
\end{table*}
\subsection{LLM summarization} 
To construct the LLM summarization baseline in our experiments, we designed prompts by combining the instructions with segment-level captions, as shown in Table~\ref{appendix_tab:llm_summ_prompt}.
This inputs guide the LLM to generate a concise and coherent video-level summary. 
\begin{table*}[t]
	\centering
	\vspace{-2mm}
	\small
	\caption{Illustration of the input construction for LLM summarization, consisting of the instructional prompt and segment-level captions.}
	\vspace{3mm}
	\label{appendix_tab:llm_summ_prompt}
	\scalebox{0.95}{
		\fbox{
			\begin{minipage}{0.9\linewidth}
			{\linespread{1.2}\selectfont
			\textbf{Instructional prompt:} \par
			Below are captions generated from individual frames of a video, each describing specific moments. Please review these frame-by-frame captions and summarize them into a single, compact caption. \par
			\par 
			\vspace{2mm}
			\textbf{Frame captions:} \par
			[1 / 6] A woman in a blue jacket is sitting in front of a sports logo. \par
			[2 / 6] Woman in blue jacket standing outdoors. \par
			[3 / 6] A man in a military uniform is standing in front of a navy sign. \par
			[4 / 6] Man in military uniform standing in front of navy sign. \par
			[5 / 6] The image shows three women wearing sports uniforms and holding medals, smiling and posing for the camera.\par
			[6 / 6] Three women wearing blue and white uniforms, smiling and holding medals. \par
			}
			\end{minipage}
		}
	}
\end{table*}

\vspace{-2mm}
\section{Failure Cases}
We present two failure cases from our framework, arising due to hallucinations in the initial segment-level captions.

\textbf{Case 1: Incorrect entity counting}  
\vspace{-2mm}
\begin{itemize}
\item \textbf{Reference captions}: [``A group of people dressed in all of the colors of the rainbow sing a happy song.'', ``Two elderly women dancing with a group of men.'', …] \vspace{-1mm}
\item \textbf{SGVC output}: ``Two guys in multi-colored tops dance in front of a wall.'' \vspace{-1mm}
\end{itemize}
While the caption accurately captures specific visual details such as ``multi-colored tops'', ``wall'', and ``dance'', the VLM hallucinate the number of individuals (``two guys'', instead of the actual group of``five people'').

\textbf{Case 2. Object misidentification}  
\vspace{-2mm}
\begin{itemize}
\item \textbf{Reference captions}: [``A man fixes a piece of machinery that appears to be a miniature tank.'', ``A guy fixing his camera equipment.'', … ] \vspace{-1mm}
\item \textbf{SGVC output}: ``A man is holding a drill in his hand while working on machinery.'' \vspace{-1mm}
\end{itemize}
The object in the person's hand is a camera, but the initial frame-level captioner incorrectly identified it as a ``drill'', influenced by the surrounding context. This hallucinated detail was propagated to the final consolidated caption.

\section{Illustration of the Overall Framework}
\label{appendix_sec:detailed_overview}
\begin{figure*}[t]
    \centering
    \includegraphics[width=0.95\textwidth]{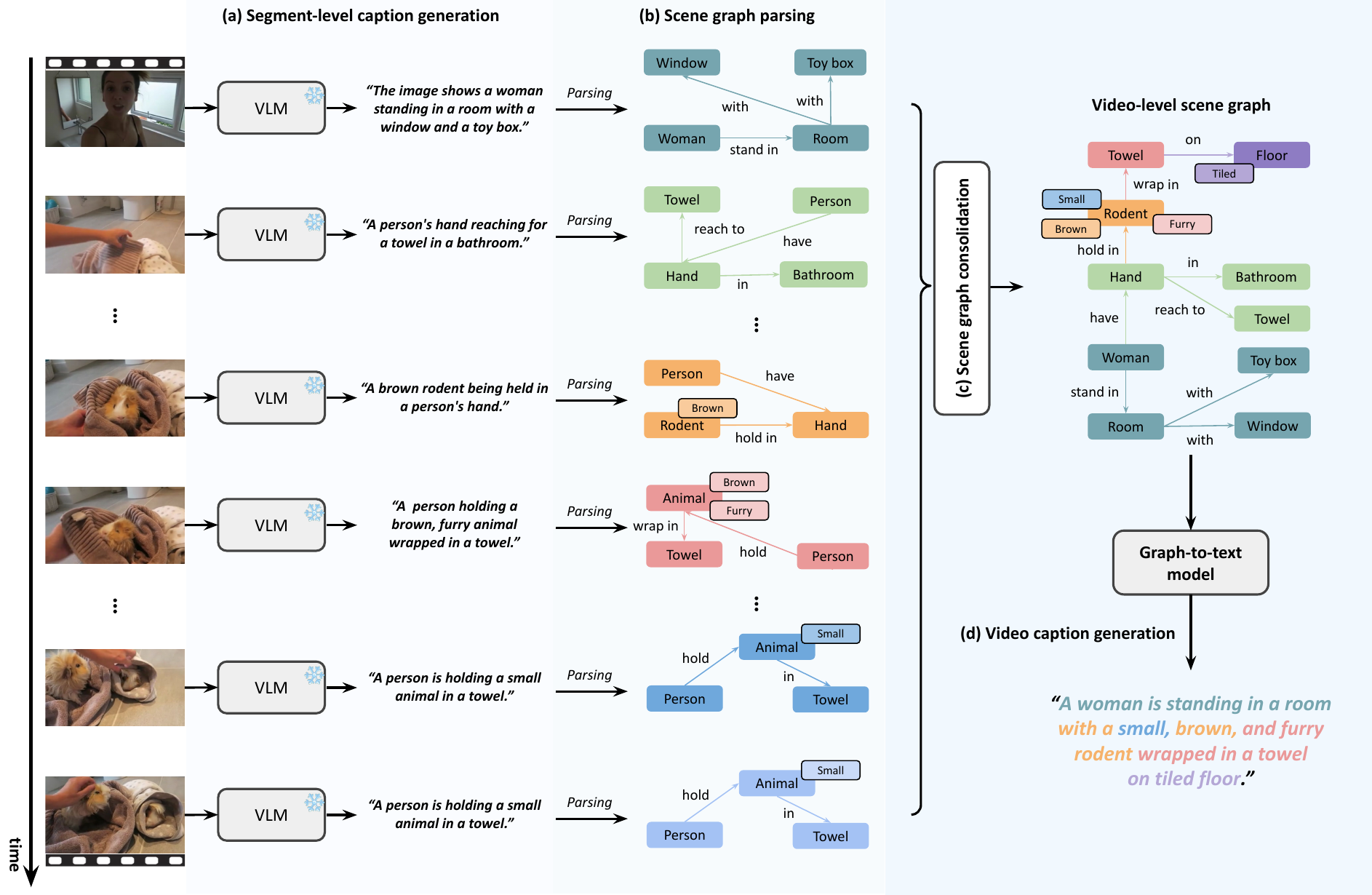} \\
     \vspace{5mm}
     \includegraphics[width=0.95\textwidth]{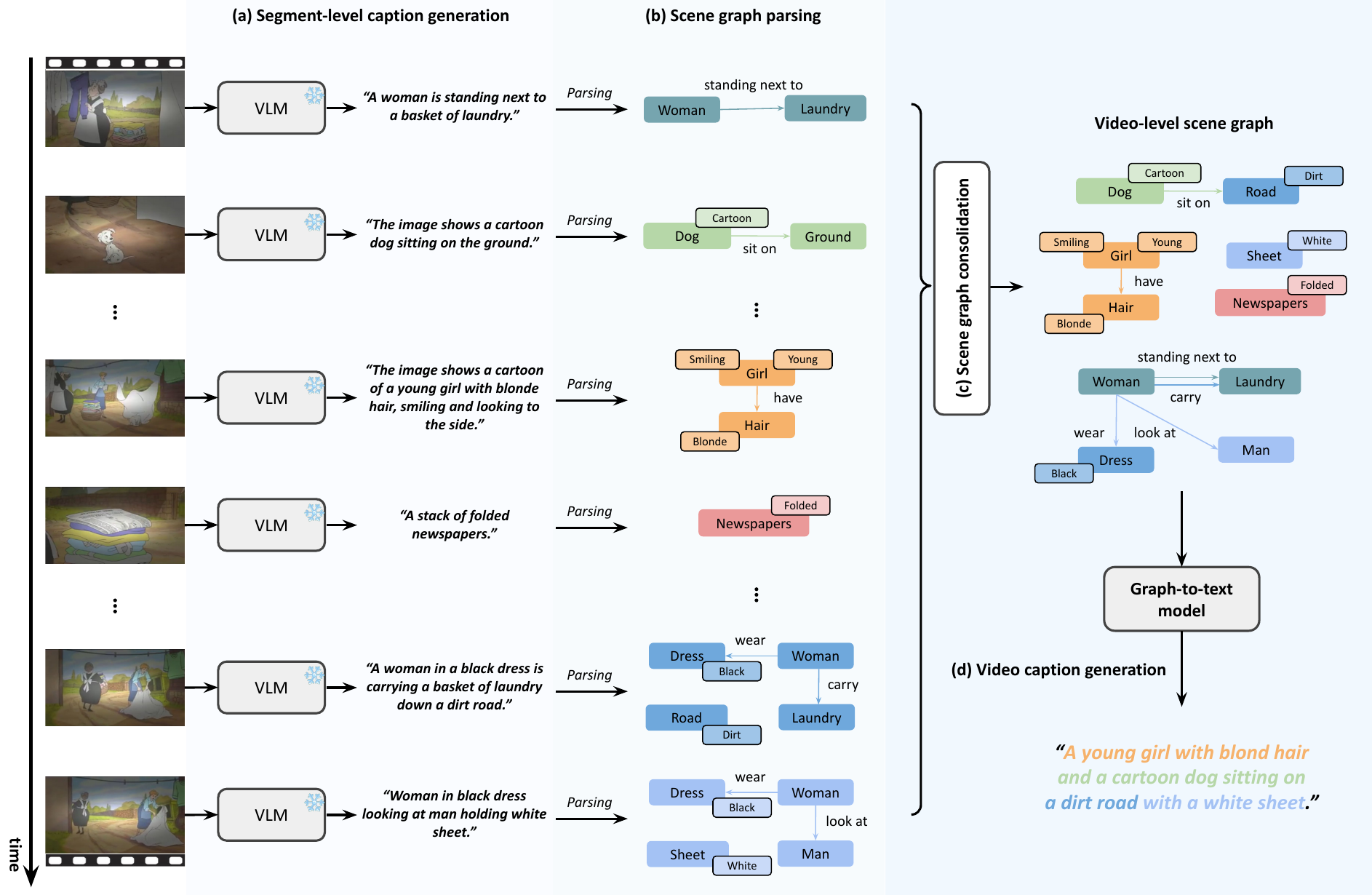}
    \caption{Illustrations of the end-to-end flow of the proposed framework. The pipeline consists of: (1) segment-level caption generation via VLMs, (2) scene graph parsing for each segments, (3) scene graph merging to produce a unified representation, and (4) graph-to-text transformation for final caption generation.}
    \label{appendix_fig:framework_detailed}
\end{figure*}

We provide illustrations of the end-to-end flow of our proposed framework for long video captioning, along with additional examples, in Figures~\ref{appendix_fig:framework_detailed}. 
The framework includes generating segment-level captions using off-the-shelf VLMs, scene graph parsing for these captions, scene graph consolidation to produce a unified representation, and graph-to-text translation for generate video generation.
\vspace{-2mm}

\section{Additional Qualitative Results}
\label{appendix_sec:qual}
\begin{figure*}[t]
    \centering
        \begin{tabular}{@{}cc@{}}
                \includegraphics[width=0.485\linewidth]{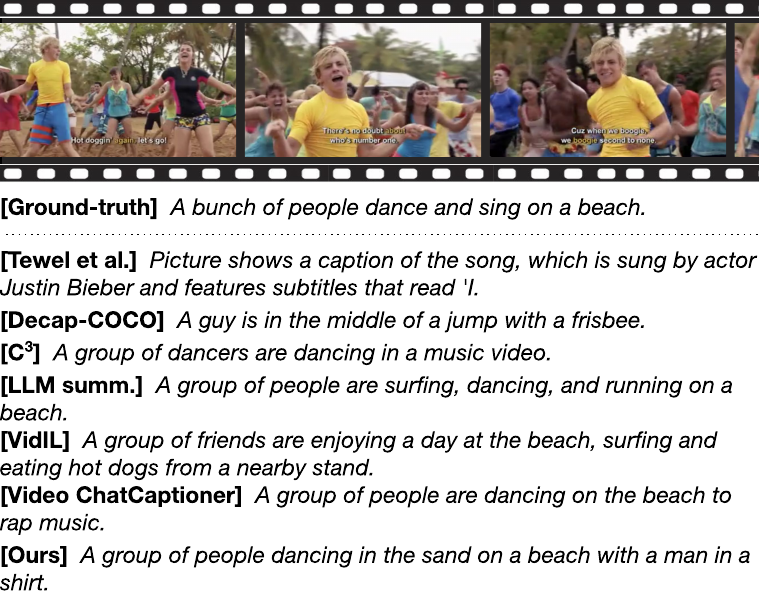} & 
                \includegraphics[width=0.485\linewidth]{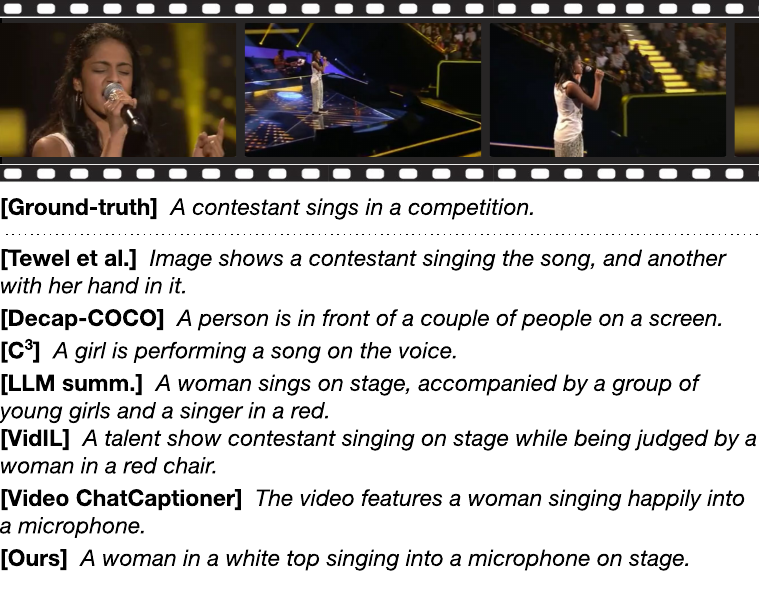} \\
                \includegraphics[width=0.485\linewidth]{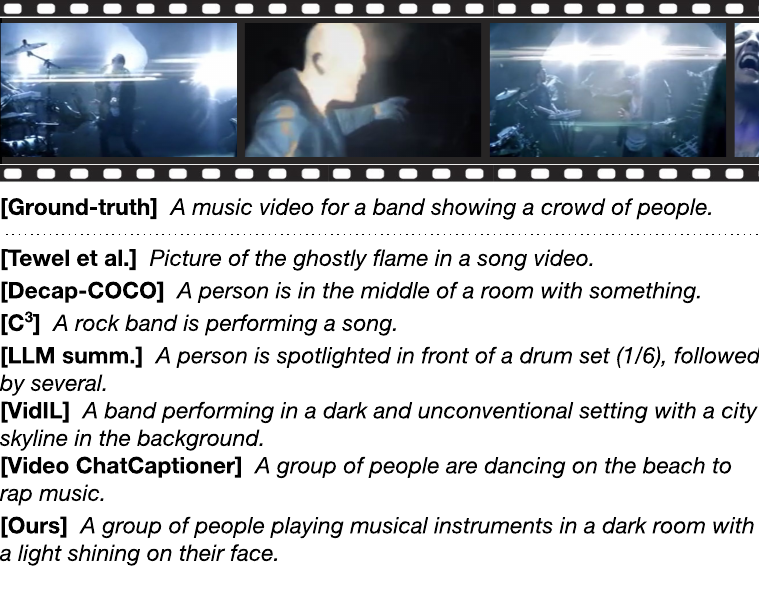} &
                \includegraphics[width=0.485\linewidth]{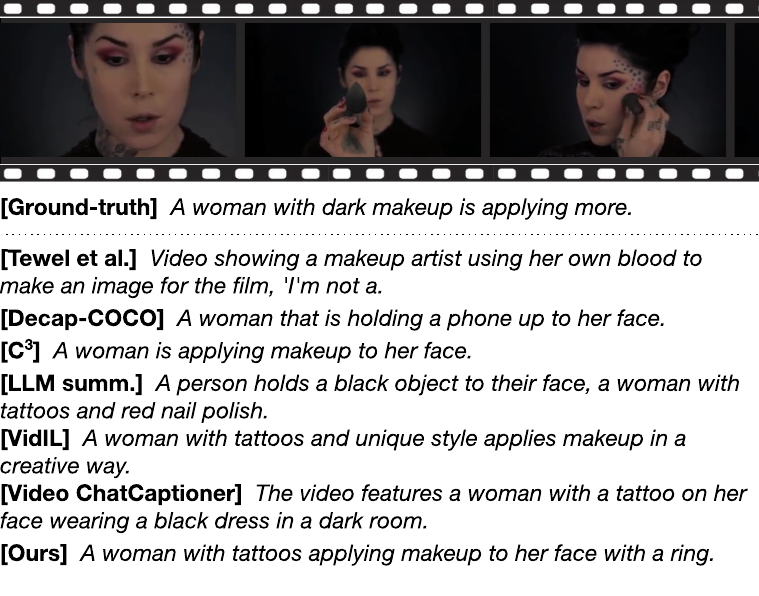} \\
                \includegraphics[width=0.485\linewidth]{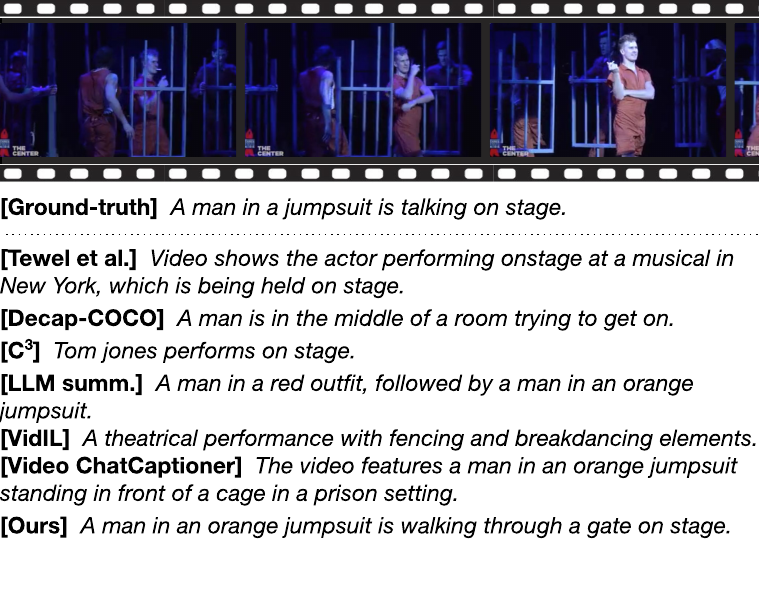} &
                \includegraphics[width=0.485\linewidth]{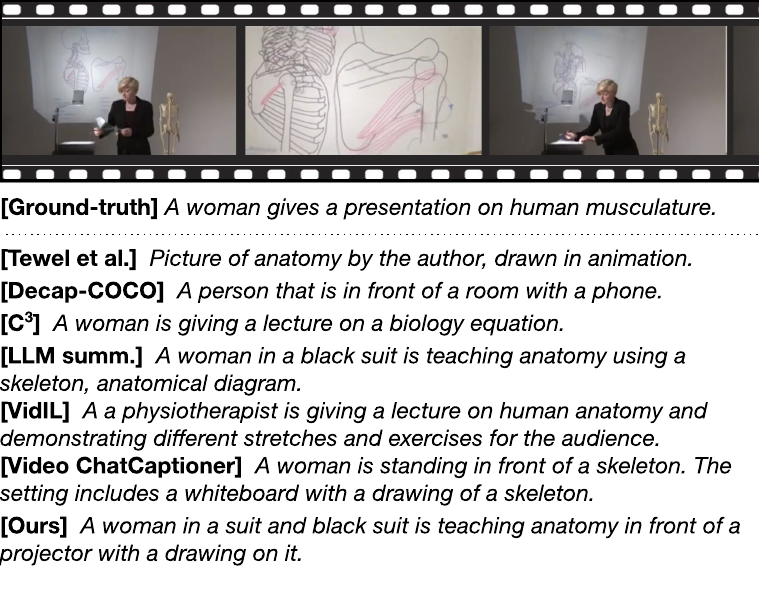} \\
        \end{tabular}
    \vspace{-2mm}
    \caption{Additional example of zero-shot video captioning results on MSR-VTT test set. We compare our results with other comparisons, listed from top to bottom as 1) Tewel \etal: test-time optimization method, 2) Decap-COCO: text-only training on COCO, 3) C$^{3}$: text-only training on MSRVTT, 4) LLM summarization using Mistral-7B-Instruct-v0.3, 5) VidIL: LLM-based video understanding with few-shot examples, 6) Video ChatCaptioner: video understanding via multi-turn conversations between VLM and LLM, and 7) SGVC (Ours).}
    \label{fig:qual_vc_supple}
\end{figure*}

\begin{figure*}[t]
    \centering
    \vspace{-10mm}
        \begin{tabular}{@{}ccc@{}}
                \includegraphics[width=0.95\linewidth]{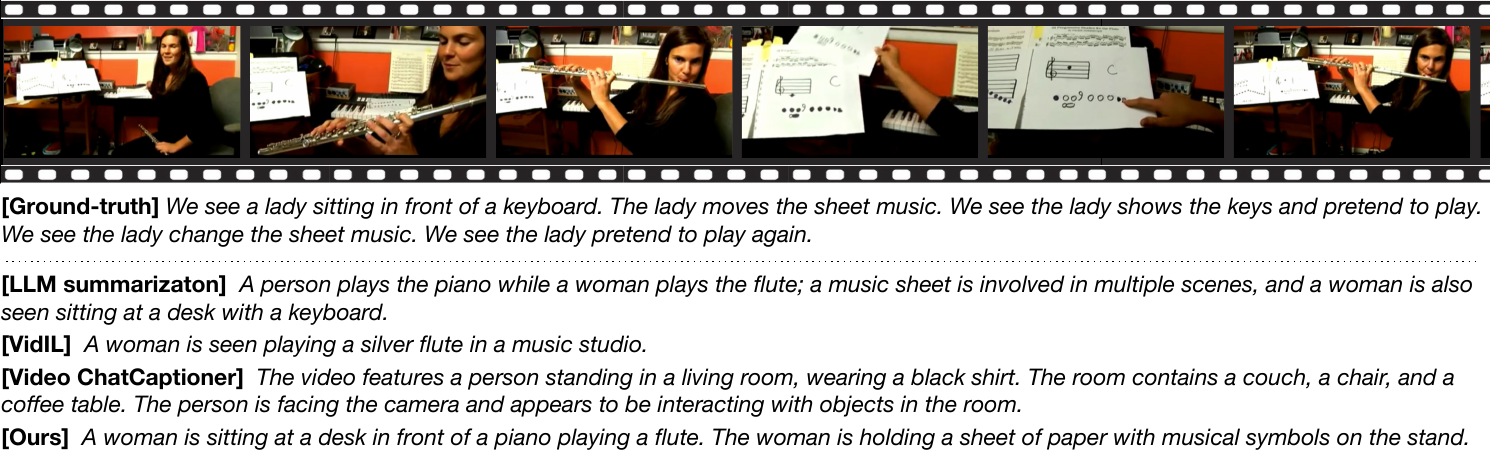} \\
                \includegraphics[width=0.95\linewidth]{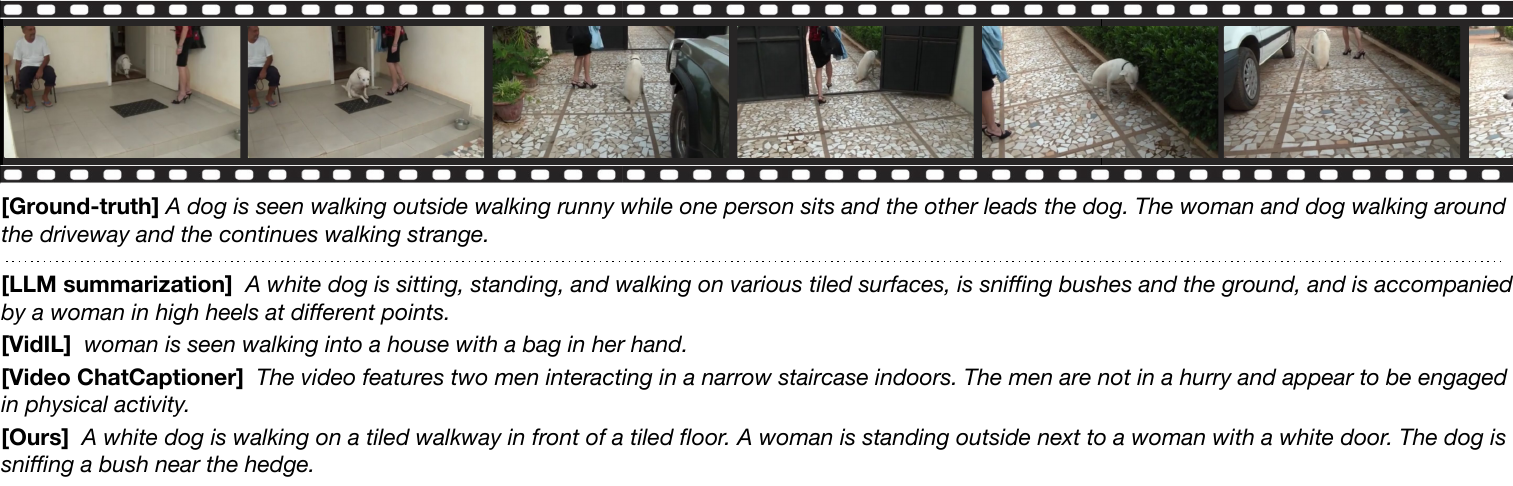} \\
                \includegraphics[width=0.95\linewidth]{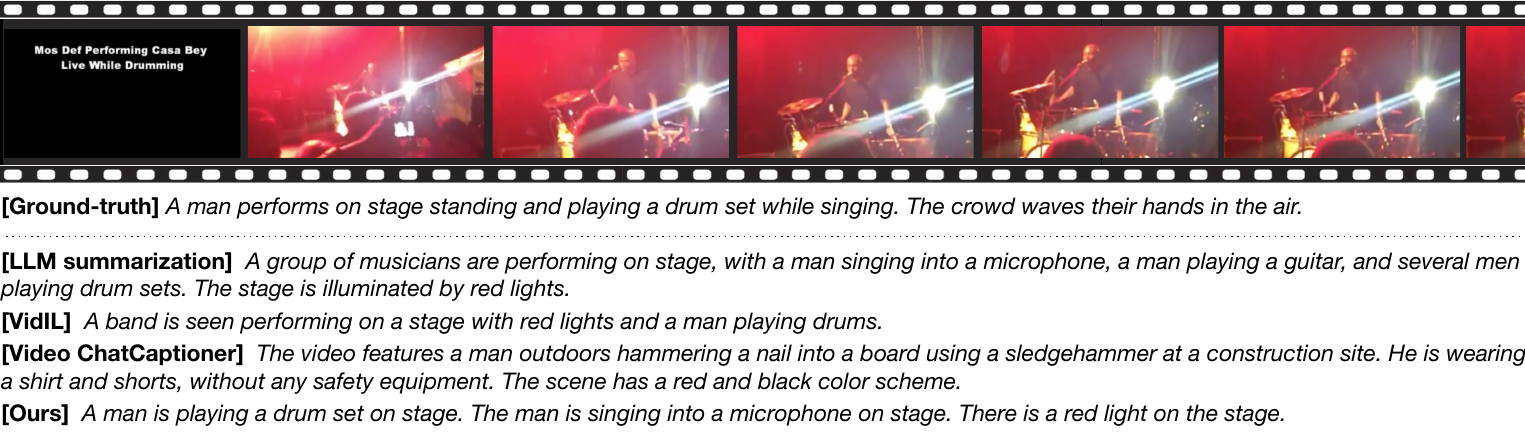} \\
                \includegraphics[width=0.95\linewidth]{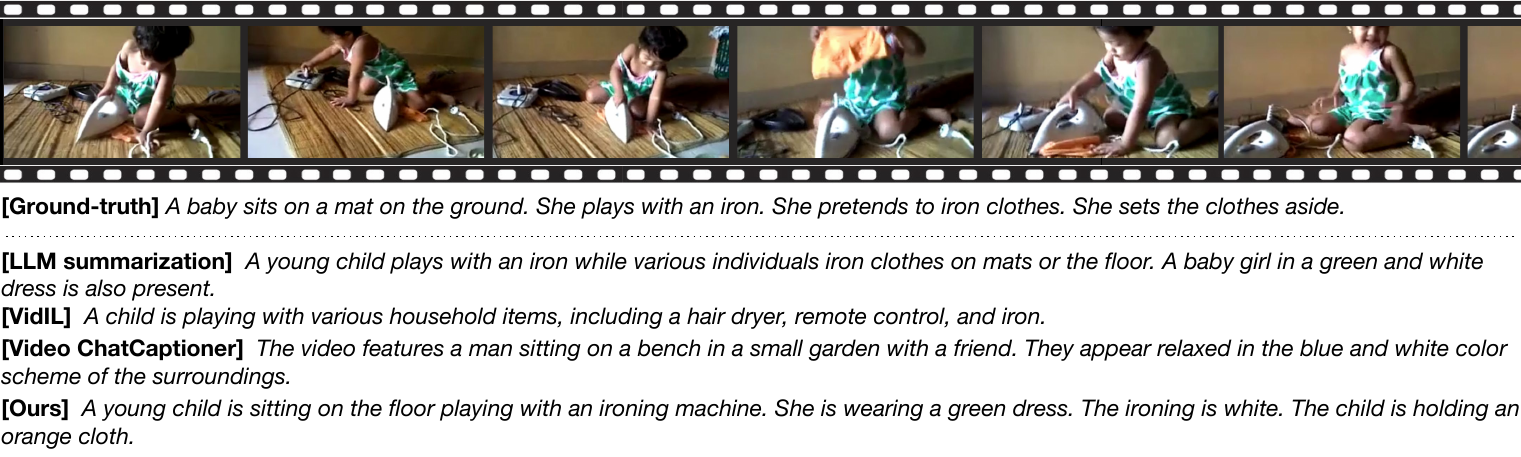} \\
            \end{tabular}
    \vspace{2mm}
    \caption{Additional example of zero-shot video paragraph captioning results on the \text{\it{ae-val}} set of the ActivityNet captions dataset. We compare our results with other comparisons, listed from top to bottom as 1) LLM summarization using Mistral-7B-instruct-v0.3, 2) VidIL, 3) Video ChatCaptioner, and 4) SGVC (Ours).}
    \label{appendix_fig:qual_vpc_supple}
\end{figure*}

We provide additional qualitative results for video captioning on the test set of MSR-VTT~\cite{xu2016msr-vtt} dataset in Figure~\ref{fig:qual_vc_supple} and for video paragraph captioning on the \text{\textit{ae-val}} set of the ActivityNet~\cite{krishna2017dense} Captions dataset in Figure~\ref{appendix_fig:qual_vpc_supple}. 


\end{document}